\documentclass[journal]{IEEEtran}
\usepackage{graphicx} 

\author{mahdi.mk24}
\date{October 2025}
\usepackage{booktabs}
\usepackage{array}
\usepackage{hyperref}
\usepackage{amsmath}
\usepackage{subfig} 
\usepackage{subcaption}
\usepackage{setspace}
\usepackage{parskip}
\usepackage{amssymb} 
\usepackage{wasysym} 
\usepackage{threeparttable} 

\usepackage{booktabs} 
\usepackage{siunitx} 
\usepackage{amssymb}  
\usepackage{pifont}   

\usepackage{amssymb}
\usepackage{algorithm}
\usepackage{algpseudocode}
\usepackage{tabularx}
\usepackage{pifont}
\usepackage{pdflscape}
\usepackage{amssymb}
\usepackage{adjustbox}
\usepackage{subfig} 
\usepackage{multirow}
\usepackage{colortbl}
\usepackage{xcolor}
\usepackage{booktabs}

\usepackage{graphicx}
\usepackage{subcaption}

\usepackage{tabularx} 

\usepackage{tabularx, xcolor, colortbl} 

\usepackage[table]{xcolor}

\usepackage{cite}
\usepackage{enumitem}
\usepackage{hyperref} 

\usepackage{amssymb} 
\usepackage{pifont}  
\usepackage{booktabs} 

\usepackage{longtable}
\usepackage{booktabs}
\usepackage{multirow}

\definecolor{best}{RGB}{198,239,206}   
\definecolor{worst}{RGB}{255,199,206}  

\title{Same Patient, Different Words, Different Diagnosis? Evaluating Semantic Stability in Clinical LLMs}

\author{Mahdi Alkaeed$^{1}$,  Adnan Qayyum$^{2}$, Nabeel Abo Kashreef$^{3}$, Muhammad Bilal$^{4}$, and Junaid Qadir$^{1}$ \\
$^1$Department of Computer Science and Engineering, College of Engineering, Qatar University, Doha, Qatar \\
$^2$College of Science and Engineering, Hamad Bin Khalifa University (HBKU), Doha, Qatar
\\
$^3$ Primary Health Care Corporation (PHCC), Doha, Qatar
\\
$^4$ Birmingham City University, Birmingham, United Kingdom.
\\
}

\begin{document}

\maketitle

\begin{abstract}
Large Language Models (LLMs) are increasingly used in clinical applications; however, their behavior remains highly sensitive to subtle linguistic variations, such as rephrasing or syntactic variation. This sensitivity poses risks in safety-critical healthcare settings, where semantically equivalent inputs should produce consistent predictions. However, a key challenge is to ensure that prompt variations truly preserve clinical meaning, as embedding-based similarity metrics often fail to capture distinctions involving negation, temporality, or severity. To address this limitation, we propose a semantic verification framework based on Natural Language Inference (NLI) to filter meaning-preserving prompt variations, which are further refined using an LLM-as-a-judge and audited by a clinical expert. In addition, we introduce three metrics to quantify model sensitivity: Meaning-Preserving Variation Sensitivity (MVS), confidence variation ($\Delta$C), and Worst-Case Instability (WCI). We evaluate 16 open-source general-purpose (GP) and medical LLMs within the same model families and parameter scales, using reformulated prompts derived from the DiagnosisQA and MedQA datasets. Our results demonstrate that robustness differences between domain-specific (DS) models are mixed and highly model-dependent, i.e., domain specialization does not consistently improve or reduce robustness to meaning-preserving prompt reformulations. Several DS models rank among the most robust (when compared with GP counterparts), and strong GP baselines remain competitive as well. We also find that confidence variation is only weakly associated with prediction instability, while overall confidence is only weakly aligned with model accuracy; many models remain overconfident despite reduced robustness or relatively low performance. This suggests that medical fine-tuning alone does not reliably guarantee stable behavior under natural prompt variations, thereby highlighting the need for explicit robustness evaluation prior to clinical deployment.
\end{abstract}


\section{Introduction}
Over the past few years, Large Language Models (LLMs) have emerged as transformative technologies that could streamline clinical workflows and enable flexible language-based interfaces to help clinicians navigate increasingly complex information landscapes. These models can support diverse tasks, including patient history summarization, discharge summary generation, medication reconciliation, and enhancing clinical diagnostics~\cite{liu2025application,subedi2025reliability}. As a result, LLMs are increasingly positioned not merely as auxiliary tools but as intermediaries between clinicians and the rapidly expanding ecosystem of structured records, unstructured clinical narratives, and evidence-based resources. 

Despite their strong performance on standardized benchmarks~\cite{singhal2023large,singhal2025toward}, recent studies reveal a critical limitation: medical LLMs exhibit significant sensitivity to minor linguistic variations in input prompts~\cite{subedi2025reliability,yan2025llm}. Such variations include lexical substitutions (e.g., \textit{myocardial infarction} vs. \textit{heart attack}), syntactic reordering, abbreviation practices, and contextual rephrasing. Ideally, clinically reliable models should exhibit semantic invariance, producing consistent outputs across meaning-preserving formulations of the same clinical scenario. However, empirical evidence suggests that performance can vary significantly under such variations, with reported deviations ranging from 8--50\% in decision-making tasks~\cite{hager2024evaluation} and up to 45\% across semantically equivalent prompt formulations~\cite{cao2024worst}. This instability poses risks to patient safety, undermines clinician trust, and complicates the deployment of LLMs in safety-critical settings.

\begin{table*}[t]
\centering
\caption{Summary of related work on semantic robustness, consistency measurement, and clinical LLM evaluation. Our work uniquely combines NLI-based semantic verification with systematic sensitivity measurement in the clinical domain.}
\label{tab:related_work}
\scriptsize
\begin{adjustbox}{max width=\textwidth}
\begin{tabular}{p{2.6cm} p{6cm} p{5.7cm} p{5.3cm}}
\hline
\textbf{Reference} & \textbf{Methodology/Dataset} & \textbf{Key Finding} & \textbf{Gap Addressed} \\
\hline

\multicolumn{4}{c}{\textit{Semantic-Preserving Perturbations}} \\
\hline
Ribeiro et al.~\cite{ribeiro2018semantically} & SEARs: Universal replacement rules via back-translation & Induced 4$\times$ more mistakes than human-discovered bugs & Lacks explicit entailment checking \\
Jin et al.~\cite{jin2020bert} & TextFooler: Synonym substitution, USE similarity $>0.84$ & Reduced BERT accuracy 5--7$\times$; 3\% changed meaning & Similarity insufficient for verification \\
Morris et al.~\cite{Morris2020} & Evaluated automated perturbations with human judgments & 38\% failed to preserve semantics; 0.98 threshold needed & Motivates NLI-based verification \\
Ross et al.~\cite{ross2022tailor} & Tailor: AMR control codes for semantic modifications & 5.8-point NLI gain from 2\% data perturbation & Demonstrates semantic control value \\

\hline
\multicolumn{4}{c}{\textit{Behavioral Testing Frameworks}} \\
\hline
Ribeiro et al.~\cite{ribeiro2020beyond} & CheckList: MFT/INV/DIR tests with templates & Users found 3$\times$ bugs; accuracy overestimates performance & Templates lack formal verification \\
Ribeiro \& Lundberg~\cite{ribeiro2022adaptive} & AdaTest: GPT-3 automated test generation & 5--10$\times$ bug-finding improvement vs. CheckList & Still lacks semantic verification \\
Wu et al.~\cite{wu2021polyjuice} & Polyjuice: GPT-2 counterfactual generator & 70\% reduction in annotation effort & Measures label-flip, not preservation \\

\hline
\multicolumn{4}{c}{\textit{Paraphrase Robustness and Consistency}} \\
\hline
Zhang et al.~\cite{Zhang2019} & PAWS: 108K pairs, human-verified (92--94.7\% agreement) & SOTA models $<40\%$ accuracy without training & Adversarial focus on lexical overlap \\
Elazar et al.~\cite{elazar2021measuring} & ParaRel: 328 paraphrases, 38 relations, manual curation & Poor consistency; high variance between relations & Manual curation, not NLI-verified \\
Verma \& Poliak~\cite{verma2021evaluating} & PaRTE: T5 paraphraser, \% $\Delta$PaRTE metric & 15--17\% prediction changes for BoW/BiLSTM & Lacks semantic verification step \\
Michail et al.~\cite{michail2025paraphrasus} & PARAPHRASUS: 43,976 pairs across 10 datasets & Best LLM: 21\% error rate & Demonstrates ongoing challenges \\

\hline
\multicolumn{4}{c}{\textit{NLI-Based Semantic Verification}} \\
\hline
Chen \& Eger~\cite{chen2023menli} & MENLI: Bidirectional NLI for generation quality & 15--30\% more robust than BERT metrics & Applied to MT/summarization, not LLM input testing \\
Laban et al.~\cite{laban2022summac} & SummaC: NLI with SummaCConv aggregation & 74.4\% accuracy; 5\% improvement on inconsistency & Document-level verification, not input equivalence \\
Mitchell et al.~\cite{mitchell2022enhancing} & ConCoRD: NLI for answer compatibility + MaxSAT & 5\% improvement on ConVQA & Verifies output consistency, not input equivalence \\
Romanov \& Shivade~\cite{romanov2018lessons} & MedNLI: 14K clinical NLI examples from MIMIC-III & Transfer learning from SNLI effective for clinical & Dataset contribution, not robustness framework \\

\hline
\multicolumn{4}{c}{\textit{Prompt Consistency and Worst-Case Performance}} \\
\hline
Cao et al.~\cite{cao2024worst} & RobustAlpacaEval: 10 paraphrases per query, manual review & Performance swings up to 45\% across formulations & General domain; manual review, not NLI verification \\
Mizrahi et al.~\cite{mizrahi2024state} & Multi-prompt LLM evaluation advocacy & Single-prompt assessments provide unstable estimates & Task-level instructions, not case-level inputs \\
Sclar et al.~\cite{sclar2023quantifying} & FormatSpread: Quantify spurious feature sensitivity & Formatting alone substantially affects performance & Tests spurious features, not semantic equivalence \\
Zhu et al.~\cite{zhu2023promptrobust} & PromptRobust: Adversarial prompt robustness & Evaluated robustness against adversarial prompts & Adversarial/degraded inputs, not semantic equivalents \\

\hline
\multicolumn{4}{c}{\textit{Clinical LLM Evaluation}} \\
\hline
Hager et al.~\cite{hager2024evaluation} & MIMIC-CDM: 2,400 cases, Llama 2 variants & 8--50\% variance from phrasing; 5--18\% from order & Clinical domain, but no systematic framework \\
Kim et al.~\cite{Kim2025} & mARC-QA: 100 adversarial clinical reasoning tasks & Best LLM 52\% vs. 66\% physicians; hallucinations & Documents limitations, not robustness framework \\
Shool et al.~\cite{shool2025systematic} & Systematic review of 761 LLM evaluation studies & 93.55\% general LLMs; 21.78\% focus on accuracy & Identifies evaluation gaps in medical domain \\
Bedi et al.~\cite{bedi2025testing} & Systematic review of 519 healthcare LLM studies & 5\% used real data; 1.2\% measured calibration & Documents lack of consistency evaluation \\

\hline
\textbf{This paper} & \textbf{NLI-verified meaning-preserving variations; MVS, $\Delta \text{C}$, worst-case instability metrics} & \textbf{First NLI-based semantic stability framework which is further refined using an LLM-
as a judge and validated by clinical experts.} & \textbf{Clinical domain; verified semantic equivalence; systematic methodology} \\
\hline
\end{tabular}
\end{adjustbox}
\end{table*}

Evaluating semantic stability requires addressing a fundamental challenge, i.e., reliably determining whether a reformulated prompt preserves the original clinical meaning. Without reliable semantic verification, it becomes impossible to distinguish genuine model instability from legitimate variation in model outputs. Existing approaches based on embedding similarity, such as cosine similarity over sentence embeddings~\cite{Reimers2019} and domain-adapted metrics such as MedSim~\cite{lei2018medsim}, are demonstrably insufficient for clinical semantic verification~\cite{you2025semantics,steck2024cosine}. These methods often fail to capture clinically critical distinctions, such as negation, temporality, or severity. For example, \textit{patient has chest pain} vs. \textit{patient denies chest pain} may receive high similarity scores despite opposite clinical meanings. 

This limitation was systematically demonstrated by Morris et al.~\cite{Morris2020}, who reported that 10\% $\sim$ 15\% of automated perturbations failed to preserve semantics even when constrained by high similarity thresholds, with 45\% introducing grammatical errors. However, even such stricter thresholds can be insufficient for clinical contexts. This leads to a methodological dilemma: to evaluate whether a clinical LLM responds consistently to meaning-preserving variations, we must first reliably identify which variations actually preserve meaning, a task for which embedding-based metrics are demonstrably inadequate. Therefore, without principled semantic verification, sensitivity analysis becomes unreliable, potentially conflating genuine instability with legitimate responses to semantic differences.

To address this limitation, we propose a Natural Language Inference (NLI)-based semantic verification framework that formulates meaning preservation as a logical entailment problem. NLI models, particularly those adapted to clinical domains (e.g., MedNLI), can explicitly determine whether a reformulated prompt \textit{entails}, \textit{contradicts}, or is \textit{neutral} with respect to the original. This framing provides a principled, interpretable mechanism for semantic verification: only variations that achieve bidirectional entailment (mutual implication) are retained for sensitivity evaluation. This ensures that the observed output variations reflect the model's genuine instability rather than legitimate responses to altered clinical content. Specifically, we attempt to answer the following key questions:

\begin{itemize}
    \item \textbf{RQ1:} Do medical LLMs outperform general-purpose models on standard medical QA benchmarks under original prompt settings (i.e., without any natural variations)?

    \item \textbf{RQ2:} To what extent does domain specialization improve prediction robustness to semantically equivalent prompt reformulations compared to general-purpose models?

    \item \textbf{RQ3:} How stable are model confidence estimates under meaning-preserving prompt reformulations, and does confidence stability align with prediction stability?

\end{itemize}


To address aforementioned RQs, we made the following salient contributions in this paper.

\begin{enumerate}
    \item We introduce an NLI-based semantic verification framework that formalizes meaning preservation as a bidirectional logical entailment, enabling sensitivity evaluation under rigorously validated semantic equivalence.
    
    \item We propose three complementary metrics: Meaning-Preserving Variation Sensitivity (MVS), Confidence Variation ($\Delta \text{C}$), and worst-case instability, which disentangle prediction variability, confidence shifts, and per-sample correctness fragility.
    
    \item We develop a semantically and human-verified benchmark of meaning-preserving prompt variations derived from two medical QA datasets and evaluate the sensitivity of both general-purpose and domain-specific LLMs.

\end{enumerate}



\section{Related Work}
\label{sec:related}
Table~\ref{tab:related_work} summarizes existing related works across four related areas: (1) clinical LLM evaluation and prompt sensitivity; (2) semantic-preserving perturbations; (3) paraphrase-based consistency measurement; and (4) NLI-based semantic verification. Below, we highlight key gaps relevant to our approach.



\subsection{Clinical LLM Evaluation and Prompt Sensitivity}
Prompt sensitivity—variation in model outputs due to minor input changes has been widely studied in the literature. For instance, in clinical settings, Hager et al.~\cite{hager2024evaluation} demonstrated that clinical LLMs exhibit 8--50\% performance variance due to phrasing and 5--18\% variance from information ordering. Similarly, Kim et al.~\cite{Kim2025} found that even the best performing LLM achieved only 52\% accuracy versus 66\% for physicians on adversarial clinical reasoning tasks, exhibiting hallucinations and overconfidence. On a similar note, studies focused on general-domain LLMs also reported similar instability. Cao et al.~\cite{cao2024worst} demonstrated performance swings of up to 45\% across semantically equivalent prompt formulations. Similarly, He et al.~\cite{he2024does} showed that prompt formatting alone can induce performance variations of up to 40\%. In a recent study, Yun et al. \cite{yun2026treatment} demonstrated that variations in question framing (e.g., positive vs. negative) can lead to contradictory model responses even when grounded in identical clinical evidence in a controlled retrieval-augmented generation (RAG) setting. 

To characterize and quantify this variability, Errica et al.~\cite{errica2025did} introduced sensitivity and consistency measures for prompt engineering, while Zhuo et al.~\cite{zhuo2024prosa} proposed the PromptSensiScore (PROSA) framework using decoding confidence to quantify response variability. In \cite{ismithdeen2025promptception}, the authors evaluated 10 large multimodal models across 61 prompt types, finding accuracy deviations up to 15\% from prompt variations. Critically, they classified prompts by instructional intent (positive/neutral/negative framing) rather than by semantic equivalence, treating prompt variability as a feature of prompt engineering rather than as a requirement for consistency. This distinction is central to our work: we focus on unintended sensitivity to meaning-preserving variations---cases where models should remain invariant but do not.


\subsection{Semantic-Preserving Perturbations and Behavioral Testing}
Early work on semantic-preserving perturbations includes Semantically Equivalent Adversarial Rules (SEARs)~\cite{ribeiro2018semantically}, which use universal replacement rules (e.g., ``What NOUN $\rightarrow$ Which NOUN'') verified through back-translation and human evaluation. Jin et al.~\cite{jin2020bert} introduced a black-box adversarial framework using synonym substitution with word importance ranking, verified by counter-fitted embeddings (cosine similarity $>0.84$) and Universal Sentence Encoder. However, human evaluation revealed that 97\% preserved semantics, indicating that 3\% still changed meaning, which highlights verification limitations. However, empirical analyses show that such constraints are insufficient; for instance, Morris et al.~\cite{Morris2020} found that 38\% of automatically generated perturbations alter meaning despite high similarity scores. This finding directly motivates our choice of NLI-based verification over similarity metrics. Similarly, behavioral testing frameworks such as CheckList~\cite{ribeiro2020beyond} and AdaTest~\cite{ribeiro2022adaptive} can systematically expose model limitations. However, they rely on templates or heuristic controls rather than formal semantic verification. As a result, they cannot guarantee that perturbations preserve meaning, limiting their applicability for consistency evaluation.

\subsection{Paraphrase Robustness and Consistency Measurement}
Several frameworks have been proposed to quantify consistency under meaning-preserving variations (i.e., paraphrases). Elazar et al. \cite{elazar2021measuring} evaluated invariance across manually curated paraphrases (with 95.5\% inter-annotator agreement) and reported high variability in model predictions despite preserved meaning. In a similar study \cite{verma2021evaluating}, the Paraphrastic Robustness in Textual Entailment (PaRTE) framework is presented that employs T5-based paraphrasers and measures changes in prediction rates. They found that bag-of-words and BiLSTM models showed 15--17\% changes in predictions, while transformer models were more robust but still changed predictions. While PaRTE provides a consistency measurement similar to our MVS metric, it lacks a semantic verification step to ensure perturbations genuinely preserve meaning.

Beyond single-prompt sensitivity, recent work examines consistency across semantically equivalent prompt formulations. Cao et al.~\cite{cao2024worst} introduced RobustAlpacaEval, a benchmark containing 10 paraphrases per query in AlpacaEval, generated with GPT-4 and manually reviewed for semantic integrity. They demonstrated performance swings up to 45\% across semantically equivalent prompts, with worst-case performance often dramatically lower than best-case performance. This worst-case analysis reveals vulnerabilities obscured by single-prompt or average-case evaluation. Mizrahi et al.~\cite{mizrahi2024state} advocated for multi-prompt evaluation, demonstrating that single-prompt assessments provide unstable estimates of model capabilities. Sclar et al.~\cite{sclar2023quantifying} introduced FormatSpread to quantify sensitivity to spurious features in prompt design and showed that formatting alone substantially affects performance. Sun et al.~\cite{sun2023evaluating} evaluated the zero-shot robustness of instruction-tuned models and found that performance varies significantly across semantically equivalent prompt reformulations.

\begin{figure*} [!t]
\centering
\includegraphics[width=0.95\linewidth]{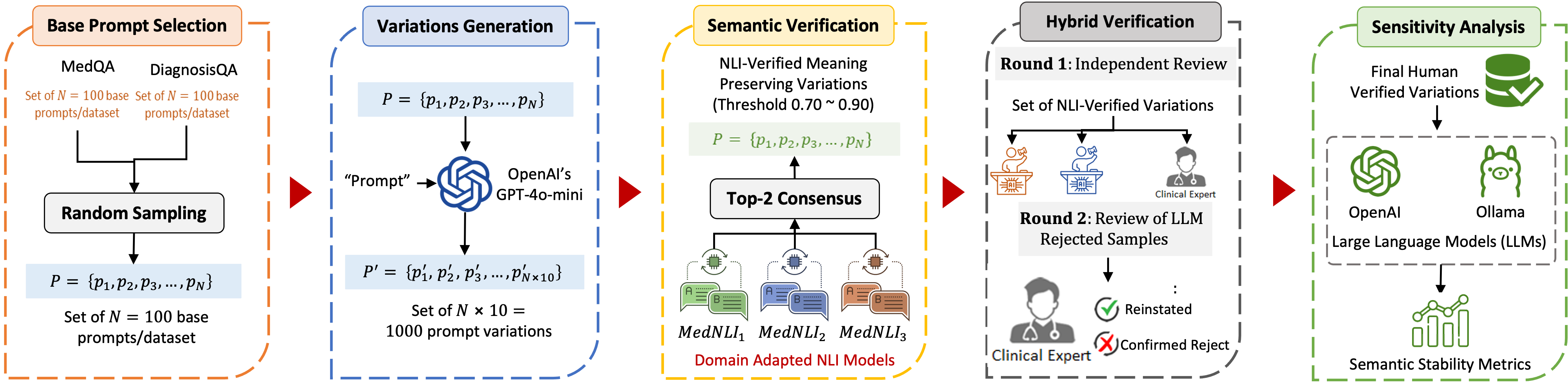}
\caption{Overview of the systematic framework for evaluating semantic
stability in clinical LLMs through meaning-preserving variations verified by Natural Language Inference models.}
\label{fig:method}
\end{figure*}

\subsection{NLI-Based Semantic Verification and Evaluation}

NLI provides a principled approach to semantic verification by framing similarity as logical entailment. Prior work has applied NLI to evaluate generation quality, detect factual inconsistencies, and enforce output-level consistency. Chen and Eger~\cite{chen2023menli} developed MENLI, using bidirectional entailment checking for text generation quality assessment. They demonstrated that NLI metrics are 15--30\% more robust to adversarial attacks than BERT-based metrics and better at detecting factual inconsistencies in machine translation and summarization. Laban et al.~\cite{laban2022summac} introduced SummaC, a method for computing NLI scores between document-summary sentence pairs using a novel SummaCConv aggregation method. SummaC achieved 74.4\% balanced accuracy (5\% improvement over prior work) on inconsistency detection, demonstrating NLI's effectiveness for document-level semantic preservation verification. 
Recent work~\cite{balamurali2025revisiting} used DeBERTa-v3 NLI models for zero-shot question-answering evaluation, matching GPT-4o accuracy (89.9\%) at orders of magnitude lower cost, demonstrating NLI's viability for evaluating semantic equivalence in question answering relevant to clinical QA scenarios.

In addition to using NLI as an evaluation metric, it has also been used to enforce consistency across model outputs. For instance, Mitchell et al.~\cite{mitchell2022enhancing} proposed ConCoRD, which uses pre-trained NLI models to check pairwise answer compatibility, builds factor graphs with NLI-based beliefs, and applies MaxSAT solvers. ConCoRD achieved 5\% improvement on ConVQA and addressed transitivity failures without fine-tuning. This work complements ours: ConCoRD uses NLI to enforce consistency across model outputs, while we use NLI to verify the semantic equivalence of inputs before measuring sensitivity. Nighojkar and  Licato~\cite{Nighojkar2021} introduced the Adversarial Paraphrasing Task (APT), which generates mutually implicative sentences with minimal lexical overlap to verify mutual implication. They highlighted the difficulty of maintaining consistency even under NLI-verified conditions. In the clinical domain, Romanov and Shivade~\cite{romanov2018lessons} created MedNLI, the first large-scale NLI dataset, comprising approximately 14,000 examples from MIMIC-III clinical notes. They demonstrated that transfer learning from SNLI is effective and that integrating medical knowledge is valuable for clinical NLI. Chowdhury et al.~\cite{chowdhury2020improving} improved MedNLI by incorporating UMLS and contextual knowledge, thereby demonstrating the value of domain-specific knowledge for clinical NLI. Our work leverages clinical NLI models, such as MedNLI, for semantic verification in clinical robustness testing, an application not explored in prior work.

\subsection{Positioning of This Work}
While prior research has advanced perturbation-based testing, consistency measurement, and NLI-based evaluation, several critical gaps remain for the reliable deployment of clinical LLMs. First, existing approaches suffer from a semantic verification gap, as they primarily rely on similarity-based constraints (e.g., cosine similarity), which fail to capture fine-grained clinical distinctions involving negation, temporality, or severity. Second, there exists a clinical consistency gap: although MedNLI~\cite{romanov2018lessons} enables clinical entailment reasoning, prior work has not systematically leveraged semantic verification to evaluate robustness under meaning-preserving variations, with most clinical evaluations focusing primarily on accuracy rather than consistency~\cite{aljohani2025comprehensive}. Third, current evaluation frameworks, such as CheckList~\cite{ribeiro2020beyond} and AdaTest~\cite{ribeiro2022adaptive}, along with consistency-focused methods like ParaRel~\cite{elazar2021measuring} and PaRTE~\cite{verma2021evaluating}, generate or assess variations without enforcing semantic equivalence, limiting their ability to isolate true model instability. Finally, a measurement and interpretation gap persists, as existing metrics quantify performance under perturbations or prompt sensitivity but do not explicitly isolate variability under meaning-preserving transformations.

\section{Methodology}
\label{sec:methodology}

In this section, we present our proposed systematic framework to evaluate semantic stability in medical LLMs under meaning-preserving linguistic variations. Figure~\ref{fig:method} illustrates the proposed pipeline, which consists of five stages: (1) base prompt selection, (2) variation generation, (3) NLI-based semantic verification, (4) hybrid judgment, and (5) sensitivity analysis. 




\subsection{Data and Variation Generation}
\subsubsection{Dataset Construction}
We construct an evaluation set of 200 base prompts that are sampled from two public medical datasets, i.e., MedQA-USMLE and DiagnosisQA~\cite{yan2025llm}. Specifically, we select 100 prompts from each dataset, representing complementary medical question answering and clinical interpretation tasks commonly used to evaluate LLMs in medical decision support settings. Each base prompt consists of a question stem, multiple-choice answer options, and a corresponding ground-truth answer. Following Cao et al. \cite{cao2024worst}, for each base prompt $p$, we generate exactly 10 candidate variations, resulting in a total of 2,000 candidate prompt variants across both datasets.

\subsubsection{Meaning-Preserving Variation Generation}

We used GPT-4o-mini with a fixed decoding temperature of 0.35 to generate ten predefined meaning-preserving variations, under strict semantic preservation constraints. These variations include abbreviation expansion, abbreviation compression, clinical-to-lay language conversion, lay-to-clinical language conversion, syntactic reordering, active-to-passive voice transformation, question reframing, documentation style modification, format shifts, and redundancy trimming. The generation process explicitly prohibits modification of clinical facts, including numeric values, units, negation, temporality, laterality, disease severity or stage, medications, and dosages. The model was further instructed not to add or remove information and to leave the stem unchanged when a requested variation could not be safely applied. To ensure diversity among generated variants, we apply a distinctness check within each prompt group using trigram Jaccard similarity, triggering regeneration when the similarity exceeded 0.92. Variants identified as identical to the original stem or insufficiently distinct from previously generated variants were regenerated up to 3 attempts per variation type. All generated variants, along with their associated metadata, were stored in a structured JSONL format for downstream processing. Table~\ref{tab:prompt_template} summarizes the prompt template used for prompt variation generation. 

\begin{table}[t]
\centering
\caption{Prompt template specifying task instructions, generation constraints, variation types, and the JSON output structure used in the meaning-preserving variations generation.}
\label{tab:prompt_template}
\scriptsize
\scalebox{0.9}{
\begin{tabular}{p{9cm}}
\toprule

\textbf{Task Context:}  
Base prompts are medical multiple-choice questions sampled from the MedQA-USMLE and DiagnosisQA datasets. Each prompt consists of a single question stem accompanied by answer options and, where available, a reference answer. Only the question stem is subject to reformulation.

\\
\midrule

\textbf{System Instruction:}  
You are a clinical documentation expert.

\\
\midrule

\textbf{Generation Instructions:}  
Rewrite \textbf{ONLY} the question stem. Do \textbf{NOT} rewrite answer options or reference answers.  
You \textbf{MUST} preserve all clinical facts exactly as given.  
Do \textbf{NOT} change negation, temporality, laterality, disease severity or stage, medications, dosages, numeric values, or units.  
Do \textbf{NOT} add facts, remove facts, or introduce uncertainty.  
If a requested variation type cannot be safely applied, keep the stem unchanged for that type and explicitly record this in \texttt{changes\_made}.  
Return \textbf{ONLY} valid JSON.

\\
\midrule

\textbf{Variation Types (exactly one per variant):}
\begin{enumerate}
    \item \texttt{abbr\_expand}
    \item \texttt{abbr\_compress}
    \item \texttt{clinical\_to\_lay}
    \item \texttt{lay\_to\_clinical}
    \item \texttt{syntax\_reorder}
    \item \texttt{active\_passive}
    \item \texttt{question\_reframe}
    \item \texttt{documentation\_style}
    \item \texttt{format\_shift}
    \item \texttt{redundancy\_trim}
\end{enumerate}

\\
\midrule

\textbf{Output Format (JSON per variant):}
{\ttfamily
\begin{tabbing}
\hspace{1cm}\=\hspace{1cm}\=\kill
\{\\
\> "type": "...",\\
\> "changes\_made": [...],\\
\> "stem": "..."\\
\}
\end{tabbing}
}

\\
\bottomrule
\end{tabular}}
\end{table}


\subsection{NLI-Based Semantic Verification}
To ensure that generated variations preserve meaning, we perform an NLI-based semantic verification procedure. This verification step operates exclusively on the question stems and is applied independently of the variation generation process. We employ three pretrained domain-specific NLI models: PubMedBERT-MNLI-MedNLI, FacebookAI's roberta-large-mnli, and Microsoft's deberta-large-mnli. Each NLI model outputs probabilities over three labels (i.e., entailment, neutral, and contradiction) for a given premise-hypothesis pair. Below, we discuss our proposed NLI-based semantic verification approach to determine whether generated prompt variations preserve the meaning of the corresponding base prompts.

\subsubsection{Bidirectional Entailment}
Let $p$ denote a base prompt and $p'$ denote a candidate variation in the set of all variations $\mathcal{V}(p)$. For each pair $(p, p')$, we compute entailment in both directions:

\vspace{-0.2mm}
\begin{equation}
\mathbf{s}_{\text{forward}} = \text{NLI}(p \rightarrow p'), \quad
\mathbf{s}_{\text{backward}} = \text{NLI}(p' \rightarrow p)
\end{equation}

Here, $\mathbf{s}_{\text{forward}} = (e_{\text{fwd}}, n_{\text{fwd}}, c_{\text{fwd}})$ and $\mathbf{s}_{\text{backward}} = (e_{\text{bwd}}, n_{\text{bwd}}, c_{\text{bwd}})$ denote the predicted probabilities for entailment, neutral, and contradiction classes in the forward and backward directions, respectively. A candidate variation is considered to satisfy bidirectional entailment for a given NLI model if the following conditions are met:
\vspace{-0.2mm}
\begin{equation}
e_{\text{fwd}} \geq \tau_{\text{ent}}, \quad
e_{\text{bwd}} \geq \tau_{\text{ent}}, \quad
c_{\text{fwd}} < \tau_{\text{con}}, \quad
c_{\text{bwd}} < \tau_{\text{con}},
\end{equation}

where $\tau_{\text{ent}}$ is the entailment threshold and $\tau_{\text{con}}$ is the contradiction threshold. This bidirectional criterion enforces mutual implication between the base prompt and its variation, reducing false positives arising from asymmetric entailment.


\subsubsection{Multi-Model Consensus}
Bidirectional entailment verification is performed independently for each of the three NLI models. For a given prompt pair $(p, p')$, a model is considered to satisfy semantic verification if it meets the predefined bidirectional threshold criteria. A multi-model consensus is then applied, requiring at least $K = 2$ of the three models to verify the pair for it to be classified as meaning-preserving. Formally, let $\mathbb{I}_m(p,p') \in {0,1}$ indicate whether model $m$ satisfies the bidirectional entailment conditions for $(p,p')$. A variation is classified as meaning-preserving if, $\sum_{m=1}^{3} \mathbb{I}_m(p,p') \geq K$. 



\subsection{Hybrid Verification Pipeline}
After prompt generation and NLI-based semantic filtering, we apply a two-round hybrid verification process to ensure that retained prompt variations preserve the original clinical meaning and context. This process consists of automated review with expert adjudication and is summarized in Table~\ref{tab:hybrid-verification-summary}. In Round 1, all NLI-verified samples are independently reviewed by both LLM-as-a-Judges and a human clinical expert. In Round 2, only samples rejected by the LLM judges are re-evaluated by the clinical expert. This adjudication step is designed to recover valid reformulations that may have been conservatively filtered by the LLM reviewers, while preserving expert oversight over ambiguous or borderline cases. The final benchmark therefore consists only of samples accepted after this sequential two-round process. Concretely, a prompt variation must (1) pass multi-model NLI semantic filtering, (2) survive independent Round 1 review, or be reinstated during Round 2 expert adjudication, and (3) satisfy final expert acceptance criteria. 

\subsubsection{LLM-as-a-Judge Evaluation and Threshold Calibration}
To complement the NLI-based verification and introduce an additional semantic filtering layer, we employ an LLM-as-a-Judge mechanism. This stage operates as a secondary validation step to assess whether candidate variations $P'$ preserve the same clinical intent as their corresponding base prompts $P$. Specifically, we employ a committee of advanced models to evaluate semantic consistency between each base prompt and its variation, utilizing a specific snapshot of \href{https://api.openai.com}{GPT-5.2} (accessed April 2026) and the \href{https://ollama.com/rjmalagon/medllama3-v20}{MedLLaMA3} model deployment via Ollama (version tag: \texttt{v20}). Unlike the NLI models, which rely on probabilistic entailment scores, the LLM-as-judge provides a binary decision (\textit{YES}/\textit{NO}) on whether semantic equivalence is preserved. Table~\ref{tab:judge_prompt_template} illustrates the prompt template used for this evaluation. 
\begin{table}[!t]
\centering
\caption{Prompt template used for validation of meaning-preserving variations using LLM-as-a-judge approach.}
\label{tab:judge_prompt_template}
\scriptsize
\scalebox{0.83}{
\begin{tabular}{p{10cm}}
\toprule

\textbf{Role:}  
You are a strict medical expert and clinical linguist.

\\
\midrule

\textbf{Task Overview:}  
Determine if the provided \textbf{Candidate Question} preserves the \textbf{EXACT} clinical meaning, intent, and factual granularity of the \textbf{Original Question}.

\\
\midrule

\textbf{Evaluation Criteria:}  
The candidate question must be rejected (NO) if there is any change in:
\begin{itemize}
    \item \textbf{Preservation of Diagnosis:} The clinical diagnosis or suspected condition must remain exactly the same. 
    \item \textbf{Preservation of Medication:} Any medications, drug names, dosages, or treatments must remain unchanged.
    \item \textbf{Preservation of Symptoms:} All patient symptoms and clinical signs must remain identical in meaning and specificity.
    \item \textbf{Preservation of Medical Records:} Any reported clinical findings, laboratory values, imaging results, vital signs, or documented history must remain unchanged.
    \item \textbf{Preservation of Overall Clinical Context:} The full clinical meaning of the case (including severity, temporality, laterality, numerical values, negations such as NOT/EXCEPT, and certainty levels) must remain consistent.
\end{itemize}

\\
\midrule

\textbf{Input Variables:}  
\textbf{Original Question:} \texttt{\{base\_stem\}} \\
\textbf{Candidate Question:} \texttt{\{candidate\_stem\}}

\\
\midrule

\textbf{Constraint:}  
 Answer: \textbf{YES} or \textbf{NO}
Reason: $<$ brief explanation if NO, otherwise write None$>$.

\\
\bottomrule
\end{tabular}}
\end{table}

\begin{figure*} [!t]
\centering
\includegraphics[width=\linewidth]{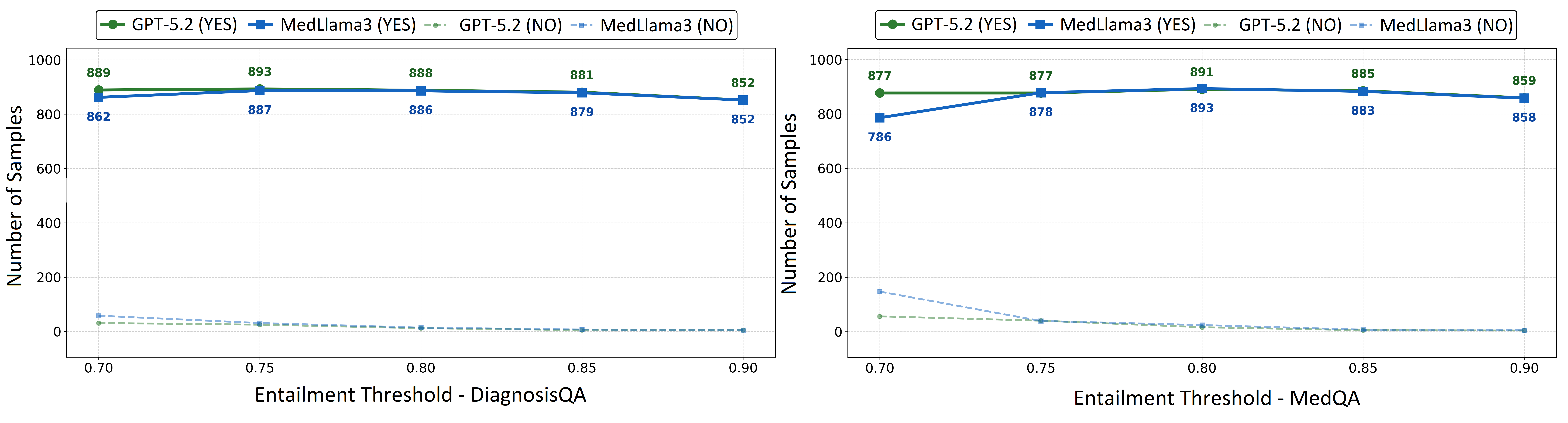}
\caption{Number of variations classified as semantically equivalent (\textit{YES}) and non-equivalent (\textit{NO}) by LLM-as-a-judge models (GPT-5.2 and MedLLaMA3) across entailment thresholds (0.70–0.90) on MedQA and DiagnosisQA datasets.}
\label{fig: Optimal Entailment Threshold}
\end{figure*}

To determine the optimal entailment threshold $\tau_{\text{ent}}$, we evaluate LLM-as-a-judge decisions across a range of thresholds from $0.70$ to $0.90$. Figure~\ref{fig: Optimal Entailment Threshold} illustrates the number of variations classified as semantically equivalent (\textit{YES}) and non-equivalent (\textit{NO}) across different thresholds for both MedQA and DiagnosisQA datasets. We define the criteria for optimal threshold based on agreement between NLI verification and LLM judgments, with the objective of maximizing accepted meaning-preserving variations while minimizing clinically inconsistent cases. As shown in Figure~\ref{fig: Optimal Entailment Threshold}, the number of accepted variations stabilizes between $0.85$ and $0.90$, indicating diminishing returns beyond this range. Therefore, we select $\tau_{\text{ent}} = 0.85$ as the operating threshold for subsequent verification.

\subsubsection{Clinical Expert Audit}
We conducted an independent clinical audit, in which one licensed clinician served as expert auditor and independently reviewed NLI-verified samples per dataset to assess whether each variation preserved the original clinical meaning and context. The evaluation was performed using the same criteria (as defined in Table~\ref{tab:judge_prompt_template}), ensuring consistency between the two LLM-based judgments and the clinical expert audit. Expert evaluation was conducted under blinded conditions, without access to LLMs' judgments, ensuring independence and minimizing potential bias. The clinical expert was conducted over three weeks to enable a careful and comprehensive assessment. The clinical expert confirmed semantic preservation with 98-99\% agreement across all samples retained after NLI-based verification, indicating that these variations consistently preserved the intended clinical meaning. This high level of agreement validates the combined filtering process. To further assess potential false negatives, the clinical expert also reviewed the samples rejected by the LLM judges. A small number of LLM-rejected samples were identified by the expert as clinically equivalent and subsequently reinstated for further analysis. Table \ref{tab:hybrid-verification-summary} summarizes the hybrid verification process to construct the meaning-preserving prompt variation benchmark.

\begin{table*}[htbp]
\centering
\caption{Summary of the hybrid verification pipeline for MedQA and DiagnosisQA datasets. In Round 1, all NLI-verified samples were independently reviewed by two LLM judges and a human expert. In Round 2, the samples rejected by the LLM judges were re-evaluated by the human expert to get final meaning-preserving variations.}
\label{tab:hybrid-verification-summary}
\scalebox{0.6}{
\begin{tabular}{lcccccccccccc}
\toprule
\multirow{3}{*}{\textbf{Dataset}} &
\multirow{3}{*}{\textbf{Base Prompts}} &
\multirow{3}{*}{\textbf{Variations}} &
\multirow{3}{*}{\textbf{NLI Verified}} &
\multicolumn{4}{c}{\textbf{Round 1 Independent Review}} &
\multicolumn{2}{c}{\textbf{Round 2 Review}} &
\multicolumn{2}{c}{\textbf{Final Dataset}} \\
\cmidrule(lr){5-8} \cmidrule(lr){9-10} \cmidrule(lr){11-12}
& & & &
\multicolumn{2}{c}{\textbf{LLM-as-a-Judge}} &
\multicolumn{2}{c}{\textbf{Human Expert}} &
\multicolumn{2}{c}{\textbf{LLM-Rejected Samples}} &
\textbf{Accepted} &
\textbf{Rejected} \\
\cmidrule(lr){5-6} \cmidrule(lr){7-8} \cmidrule(lr){9-10}
& & & &
\textbf{Accepted} &
\textbf{Rejected} &
\textbf{Accepted} &
\textbf{Rejected} &
\textbf{Reinstated} &
\textbf{Confirmed Reject} &
&
\\
\midrule
MedQA       & 100 & 1000 & 890  & 878  & 12 & 883  & 7  & 5 & 7  & 883  & 7  \\
DiagnosisQA & 100 & 1000 & 886  & 875  & 11 & 878  & 8  & 3 & 8  & 878  & 8  \\
\midrule
Total       & 200 & 2000 & 1776 & 1753 & 23 & 1761 & 15 & 8 & 15 & 1761 & 15 \\
\bottomrule
\end{tabular}}
\end{table*}

\subsection{Semantic Stability Metrics}
We propose a set of quantitative metrics to measure the stability of model outputs under meaning-preserving prompt variations. These metrics quantify different aspects of stability, including prediction consistency, confidence variation, and worst-case robustness under semantically equivalent inputs. The metrics are defined independently of any specific model architecture or task and can be applied to both classification and generation settings. Let $p$ denote a base prompt, and $\mathcal{V}(p)$ denote the set of verified meaning-preserving variations associated with $p$. Let $\mathcal{M}$ denote a model under evaluation.

\subsubsection{Meaning-Preserving Variation Sensitivity (MVS)}
It quantifies the proportion of verified variations that lead to a change in the model output relative to the base prompt, defined as:


\begin{equation}
\text{MVS}_{\mathcal{M}}(p) = \frac{1}{|\mathcal{V}(p)|} \sum_{p' \in \mathcal{V}(p)} \mathbf{1}\left[\mathcal{M}(p) \neq \mathcal{M}(p')\right]
\end{equation}


For classification tasks, $\mathcal{M}(p) \neq \mathcal{M}(p')$ indicates that the predicted class differs between the base prompt and its variation. For generation tasks, this condition can be instantiated using task-specific equivalence criteria (e.g., exact match or semantic equivalence). Lower values of MVS indicate greater semantic stability, with $\text{MVS} = 0$ corresponding to perfect invariance.

\subsubsection{Confidence Variation ($\Delta \text{C}$)}
It captures changes in model confidence across meaning-preserving prompt variations, even when predicted outputs remain unchanged. Let $C_{\mathcal{M}}(p)$ denote the maximum predicted probability associated with the model output for prompt $p$,  $\Delta \text{C}$ is defined as:

\begin{equation}
\Delta C_{\mathcal{M}}(p) = \frac{1}{|\mathcal{V}(p)|} \sum_{p' \in \mathcal{V}(p)} \left| C_{\mathcal{M}}(p) - C_{\mathcal{M}}(p') \right|
\end{equation}

Lower values of $\Delta C_{\mathcal{M}}(p)$ indicate more stable confidence behavior across meaning-preserving variations.

\subsubsection{Worst-Case Per-Sample Instability (WCI)}
While MVS quantifies how frequently model predictions change under meaning-preserving variations, it does not directly capture whether such changes affect answer correctness. To address this, we introduce a worst-case per-sample instability metric that measures whether the model's correctness is invariant under all verified variations of a prompt. Intuitively, this metric checks whether any meaning-preserving rephrasing can flip the model from correct to incorrect (or vice versa). Let $y(p)$ denote the ground-truth label associated with the prompt $p$, 
\[
s_{\mathcal{M}}(p) = \mathbf{1}\left[\mathcal{M}(p) = y(p)\right]
\]
denote the binary correctness of model $\mathcal{M}$ on $p$. We define the worst-case correctness over the set $\{p\} \cup \mathcal{V}(p)$ as:
\begin{align}
s_{\mathcal{M}}^{\text{best}}(p) &= \max\Big( s_{\mathcal{M}}(p), \max_{p' \in \mathcal{V}(p)} s_{\mathcal{M}}(p') \Big), \\
s_{\mathcal{M}}^{\text{worst}}(p) &= \min\Big( s_{\mathcal{M}}(p), \min_{p' \in \mathcal{V}(p)} s_{\mathcal{M}}(p') \Big).
\end{align}

The worst-case per-sample instability is then defined as:
\begin{equation}
\text{Instability}_{\mathcal{M}}(p) =
s_{\mathcal{M}}^{\text{best}}(p) - s_{\mathcal{M}}^{\text{worst}}(p)
\end{equation}

This metric takes values in $\{0,1\}$. A value of $0$ indicates that the model's correctness is invariant across all meaning-preserving variations of $p$, while a value of $1$ indicates that at least one variation causes a change in correctness. Lower values, therefore, correspond to stronger worst-case robustness.

\section{Experimental Setup}

\subsection{Evaluation Protocol}
All experiments are conducted on prompt variations that satisfy the multi-stage semantic verification pipeline, including NLI-based verification, LLM-as-a-judge filtering, and clinical expert audit. For each verified sample, we compute model predictions for both the original base prompt and all associated semantically-verified variations, facilitating direct comparison under controlled semantic conditions. We evaluate model sensitivity in a multiple-choice QA setting, where each input consists of a clinical question stem and a fixed set of answer options. The order of answer options remains unchanged across all prompt variations, ensuring that observed differences in model behavior are attributable solely to the linguistic reformulation of the question stem and isolating semantic sensitivity from confounding factors such as answer space variation or task ambiguity.

\subsection{Model Selection}
We evaluate a diverse set of open-weight large language models spanning both general-purpose (GP) and medically adapted domain-specific (DS) variants across similar families, with comparable parameter sizes ranging from 1.5B to 8B. The GP models include Qwen2.5-1.5B, Qwen2.5-3B, LLaMA-3.2-3B, Qwen2.5-7B, Mistral-7B-v0.2, LLaMA3.1-8B, and Qwen3-8B. These models are trained primarily on broad-domain corpora and serve as competitive baselines for reasoning and language understanding. The DS models include MedQwen2.5-1.5B, MedQwen2.5-3B, MedLLaMA-3.2-3B, BioMistral-7B, MedQwen2.5-7B, Meditron3-8B, MedLLaMA3-8B, MediChat-LLaMA3-8B, and Qwen3-8B-Biomedical. These models incorporate biomedical or clinical training data and are designed to better capture medical terminology and DS reasoning patterns. Including both GP and DS models enables us to assess whether domain specialization improves robustness to meaning-preserving linguistic reformulations.

\subsection{Model Inference and Configuration Settings}
All models are executed locally using the Hugging Face \texttt{transformers} library in inference-only mode, ensuring a consistent evaluation across different models. To enable efficient evaluation, we perform inference using 4-bit NF4 quantization with half-precision computation, which substantially reduces memory requirements while maintaining practical inference fidelity for comparative analysis. Moreover, to ensure consistency across all models, we use a standardized prompt template that instructs each model to act as a careful medical reasoning assistant, analyze the medical question, and select exactly one answer option from A--E. The prompt includes the question stem, all five answer choices, and concludes with the expected output cue \texttt{Answer:}. 

Rather than relying on unconstrained free-form text generation, we derive predictions directly from the model's next-token output distribution following the \texttt{Answer:} cue. The candidate answer space is restricted to the five option tokens (A--E), and the selected prediction corresponds to the option receiving the highest probability. Because predictions are obtained directly from token probabilities rather than stochastic decoding, the evaluation procedure is fully deterministic and unaffected by sampling-based generation settings. We compute confidence by normalizing the probabilities assigned to the five answer-option tokens and selecting the probability of the predicted option (i.e., the highest value). Therefore, the reported confidence score reflects the model's relative preference among the available answer choices rather than a fully calibrated probability of correctness.


\section{Results and Discussions}
\label{sec:results}


\begin{table}[t]
\centering
\caption{Comparison of GP and DS models on DiagnosisQA and MedQA, with best (green) and worst (red) results highlighted by dataset and model type.}
\label{tab:baseline_results}
\resizebox{\columnwidth}{!}{%
\begin{tabular}{lllcccc}
\toprule
& & & \multicolumn{2}{c}{\textbf{DiagnosisQA}} & \multicolumn{2}{c}{\textbf{MedQA}} \\
\cmidrule(lr){4-5} \cmidrule(lr){6-7}
\textbf{Model} & \textbf{Type} & \textbf{Params} & \textbf{Acc} & \textbf{Conf} & \textbf{Acc} & \textbf{Conf} \\
\midrule

Qwen 2.5        & GP & 1.5B 
& \cellcolor{worst}$0.46$
& \cellcolor{worst}$0.46 \pm 0.13$
& \cellcolor{worst}$0.29$
& \cellcolor{worst}$0.42 \pm 0.10$ \\

LLaMA 3.2       & GP & 3B   
& $0.77$
& $0.64 \pm 0.19$
& $0.54$
& $0.51 \pm 0.17$ \\

Qwen 2.5        & GP & 3B   
& $0.64$
& \cellcolor{best}$0.95 \pm 0.10$
& $0.40$
& \cellcolor{best}$0.90 \pm 0.16$ \\

Mistral         & GP & 7B   
& $0.53$
& $0.91 \pm 0.16$
& $0.38$
& \cellcolor{best}$0.91 \pm 0.16$ \\

Qwen 2.5        & GP & 7B   
& $0.73$
& $0.90 \pm 0.16$
& $0.49$
& $0.89 \pm 0.15$ \\

LLaMA 3.1       & GP & 8B   
& \cellcolor{best}$0.80$
& $0.80 \pm 0.18$
& \cellcolor{best}$0.66$
& $0.62 \pm 0.23$ \\

Qwen 3          & GP & 8B   
& $0.73$
& $0.76 \pm 0.19$
& $0.56$
& $0.69 \pm 0.21$ \\

\midrule

MedQwen 2.5     & DS & 1.5B 
& $0.42$
& \cellcolor{worst}$0.44 \pm 0.12$
& $0.29$
& \cellcolor{worst}$0.41 \pm 0.10$ \\

MedLLaMA 3.2    & DS & 3B   
& $0.75$
& $0.76 \pm 0.21$
& $0.55$
& $0.58 \pm 0.21$ \\

MedQwen 2.5     & DS & 3B   
& $0.64$
& \cellcolor{best}$0.95 \pm 0.10$
& $0.40$
& \cellcolor{best}$0.90 \pm 0.16$ \\


BioMistral      & DS & 7B   
& $0.51$
& $0.62 \pm 0.20$
& $0.36$
& $0.58 \pm 0.20$ \\

MedQwen 2.5     & DS & 7B   
& $0.79$
& $0.78 \pm 0.20$
& $0.52$
& $0.60 \pm 0.22$ \\

MedLLaMA 3      & DS & 8B   
& $0.69$
& $0.80 \pm 0.21$
& $0.51$
& $0.70 \pm 0.22$ \\

MediChat-LLaMA 3& DS & 8B   
& $0.59$
& $0.75 \pm 0.21$
& $0.44$
& $0.68 \pm 0.19$ \\

Meditron 3      & DS & 8B   
& \cellcolor{best}$0.80$
& $0.68 \pm 0.19$
& $0.59$
& $0.50 \pm 0.19$ \\

Qwen 3 Biomedical & DS & 8B   
& $0.77$
& \cellcolor{best}$0.84 \pm 0.17$
& \cellcolor{best}$0.72$
& \cellcolor{best}$0.73 \pm 0.21$ \\

\bottomrule
\end{tabular}%
}
\end{table}

\begin{table*}[t]
\centering
\caption{Sensitivity analysis of GP and DS LLMs on DiagnosisQA and MedQA datasets. For each dataset and model type, the best (lowest) and worst (highest) values, are highlighted in green and red, respectively.}
\label{tab:sensitivity_analysis}
\scalebox{0.75}{
\begin{tabular}{l c ccc ccc}
\toprule
\textbf{Model} & \textbf{Params} &
\multicolumn{3}{c}{\textbf{DiagnosisQA}} &
\multicolumn{3}{c}{\textbf{MedQA}} \\
\cmidrule(lr){3-5} \cmidrule(lr){6-8}
& &
\textbf{MVS} ($\downarrow$) & $\Delta$C ($\downarrow$) & \textbf{WCI} ($\downarrow$) &
\textbf{MVS} ($\downarrow$) & $\Delta$C ($\downarrow$) & \textbf{WCI} ($\downarrow$) \\
\midrule

\multicolumn{8}{c}{\textbf{General-Purpose Models (GP)}} \\
\midrule

Qwen2.5-1.5B & 1.5B &
0.13 $\pm$ 0.21 & \cellcolor{best}0.04 $\pm$ 0.02 & 25.25\% &
0.11 $\pm$ 0.21 & \cellcolor{best}0.04 $\pm$ 0.02 & \cellcolor{best}15.00\% \\
\cmidrule(lr){1-8}

Qwen2.5-3B & 3B &
0.13 $\pm$ 0.21 & 0.05 $\pm$ 0.07 & \cellcolor{worst}30.30\% &
\cellcolor{worst}0.17 $\pm$ 0.25 & 0.07 $\pm$ 0.08 & \cellcolor{worst}36.00\% \\

LLaMA-3.2-3B & 3B &
0.08 $\pm$ 0.17 & 0.05 $\pm$ 0.02 & 25.25\% &
0.13 $\pm$ 0.21 & 0.06 $\pm$ 0.03 & 33.00\% \\
\cmidrule(lr){1-8}

Qwen2.5-7B & 7B &
0.11 $\pm$ 0.21 & \cellcolor{worst}0.07 $\pm$ 0.09 & \cellcolor{worst}30.30\% &
\cellcolor{best}0.07 $\pm$ 0.15 & 0.07 $\pm$ 0.07 & 18.00\% \\

Mistral-7B-v0.2 & 7B &
\cellcolor{worst}0.16 $\pm$ 0.26 & 0.06 $\pm$ 0.08 & 28.28\% &
0.15 $\pm$ 0.24 & \cellcolor{worst}0.08 $\pm$ 0.09 & 27.00\% \\
\cmidrule(lr){1-8}

LLaMA3.1-8B & 8B &
\cellcolor{best}0.07 $\pm$ 0.16 & 0.06 $\pm$ 0.04 & \cellcolor{best}20.20\% &
0.12 $\pm$ 0.24 & 0.06 $\pm$ 0.04 & 26.00\% \\

Qwen3-8B & 8B &
0.10 $\pm$ 0.19 & \cellcolor{worst}0.07 $\pm$ 0.05 & 28.28\% &
0.12 $\pm$ 0.20 & 0.07 $\pm$ 0.05 & 32.00\% \\

\midrule
\multicolumn{8}{c}{\textbf{Domain-Specific Models (DS)}} \\
\midrule

MedQwen2.5-1.5B & 1.5B &
\cellcolor{worst}0.14 $\pm$ 0.23 & \cellcolor{best}0.04 $\pm$ 0.02 & 26.26\% &
0.14 $\pm$ 0.23 & \cellcolor{best}0.04 $\pm$ 0.02 & \cellcolor{best}20.00\% \\
\cmidrule(lr){1-8}

MedQwen2.5-3B & 3B &
0.13 $\pm$ 0.21 & 0.05 $\pm$ 0.07 & 30.30\% &
\cellcolor{worst}0.17 $\pm$ 0.24 & 0.07 $\pm$ 0.09 & 36.00\% \\

MedLLaMA-3.2-3B & 3B &
0.10 $\pm$ 0.19 & 0.06 $\pm$ 0.04 & 30.30\% &
0.15 $\pm$ 0.24 & 0.07 $\pm$ 0.05 & 32.00\% \\
\cmidrule(lr){1-8}

BioMistral-7B & 7B &
\cellcolor{worst}0.14 $\pm$ 0.22 & \cellcolor{worst}0.07 $\pm$ 0.05 & \cellcolor{worst}34.34\% &
0.14 $\pm$ 0.22 & 0.06 $\pm$ 0.03 & 28.00\% \\

MedQwen2.5-7B & 7B &
0.10 $\pm$ 0.20 & 0.05 $\pm$ 0.05 & 30.30\% &
\cellcolor{best}0.13 $\pm$ 0.19 & 0.06 $\pm$ 0.04 & 32.00\% \\
\cmidrule(lr){1-8}

Meditron3-8B & 8B &
\cellcolor{best}0.09 $\pm$ 0.20 & 0.05 $\pm$ 0.03 & 21.21\% &
0.16 $\pm$ 0.27 & 0.05 $\pm$ 0.03 & 30.00\% \\

MedLLaMA3-8B & 8B &
\cellcolor{best}0.09 $\pm$ 0.17 & 0.06 $\pm$ 0.05 & 24.24\% &
0.16 $\pm$ 0.26 & 0.08 $\pm$ 0.06 & 29.00\% \\

MediChat-LLaMA3-8B & 8B &
0.11 $\pm$ 0.22 & \cellcolor{worst}0.07 $\pm$ 0.05 & 23.23\% &
0.16 $\pm$ 0.24 & \cellcolor{worst}0.09 $\pm$ 0.06 & 27.00\% \\

Qwen3-8B-Biomedical & 8B &
\cellcolor{best}0.09 $\pm$ 0.19 & 0.05 $\pm$ 0.05 & \cellcolor{best}18.18\% &
0.15 $\pm$ 0.24 & 0.07 $\pm$ 0.05 & \cellcolor{worst}37.00\% \\

\bottomrule
\end{tabular}
}
\end{table*}

\subsection{Baseline GP vs. DS Performance (RQ1)}
Before evaluating sensitivity to meaning-preserving prompt variations, we establish a baseline using the original prompts from DiagnosisQA and MedQA. This baseline serves two purposes: it quantifies each model's nominal capability under standard prompting conditions, and it provides a reference point for interpreting behavioral changes under prompt reformulations. Table~\ref{tab:baseline_results} reports baseline accuracy and confidence scores for GP and DS LLMs across comparable parameter scales (1.5B--8B) and similar model families. It is evident from that table that \textit{DS LLMs do not consistently outperform GP models under original prompt settings}. While several DS models achieve clear gains over their corresponding GP counterparts, these improvements are selective rather than universal. On DiagnosisQA, the strongest GP model (LLaMA 3.1 8B, 0.80) matches the best DS model (Meditron 3 8B, 0.80), and also exceeds several medically specialized alternatives. On MedQA, GP models remain competitive, with LLaMA 3.1 8B reaching 0.66 accuracy, surpassed only by Qwen 3 Biomedical (0.72). These results show that strong GP instruction-tuned models can perform competitively on medical QA benchmarks without explicit domain specialization.


Table~\ref{tab:baseline_results} also shows that domain specialization can provide meaningful benefits when the adaptation process is effective. For instance, MedQwen 2.5 7B improves over Qwen 2.5 7B on both DiagnosisQA (0.79 vs. 0.73) and MedQA (0.52 vs. 0.49), while Qwen 3 Biomedical substantially exceeds standard Qwen 3 8B, especially on MedQA (0.72 vs. 0.56). These improvements suggest that targeted biomedical adaptation can enhance domain knowledge and reasoning. Notably, the strongest performance on MedQA is achieved by the Qwen 3 Biomedical (DS model), showing that well-executed domain adaptation can yield substantial gains. However, this effect is not seen uniformly across different DS models, such as BioMistral and MediChat-LLaMA 3, which underperform similarly sized GP baselines. In particular, the sizeable gap between MediChat-LLaMA 3 (0.59) and its stronger GP counterpart LLaMA 3.1 8B (0.80) on DiagnosisQA may reflect catastrophic forgetting, where domain-specific fine-tuning partially degrades broader reasoning capabilities learned during base pretraining. This suggests that the impact of specialization depends more on how adaptation is performed, including data quality, objective design, alignment strategy, and retention of general capabilities rather than on whether medical adaptation.

We also observe a persistent gap between predictive accuracy and model confidence across datasets. Nearly all models achieve lower accuracy on MedQA than on DiagnosisQA, suggesting that MedQA is the more challenging benchmark and likely requires deeper factual recall or more complex reasoning. Despite this drop in performance, confidence scores remain high for several models across both datasets, often ranging from 0.70 to 0.95, even when accuracy is modest. For example, Qwen 2.5 3B and MedQwen 2.5 3B achieve only 40\% accuracy on MedQA while reporting confidence of 0.90, and Mistral 7B provides just 38\% accuracy with confidence as high as 0.91. This misalignment indicates systematic overconfidence, where models express strong certainty without corresponding reliability. Importantly, this behavior occurs in both GP and DS models, suggesting that medical fine-tuning alone does not consistently resolve confidence misalignment.


\subsection{Analyzing GP and DS LLMs Under Meaning-Preserving Variations (RQ2)}

\subsubsection{Aggregate Sensitivity Analysis}
To answer RQ2, we first provide an aggregate sensitivity analysis in Table~\ref{tab:sensitivity_analysis}, which analyzes whether domain specialization consistently improves robustness when prompts are reformulated without changing their meaning and context. These results demonstrate that domain specialization does not provide a universal robustness advantage over GP models, but neither does it systematically reduce robustness. Instead, its effects are selective, model-specific, and dataset-dependent. Across both DiagnosisQA and MedQA, several DS LLMs rank among the most stable models, while strong GP models remain highly competitive and frequently match DS counterparts. For example, on DiagnosisQA, LLaMA3.1-8B provides the lowest MVS among GP models (0.07) with a relatively low WCI (20.20\%), while Qwen2.5-7B performs strongly on MedQA, with the lowest GP MVS (0.07) and a low WCI (18.00\%). Among DS models, Qwen3-8B-Biomedical achieves the lowest WCI on DiagnosisQA (18.18\%), and Meditron3-8B and MedLLaMA3-8B also demonstrate consistently strong robustness. On MedQA, MedQwen2.5-7B provides the lowest DS MVS (0.13), while MedQwen2.5-1.5B attains the lowest DS WCI (20.00\%). Therefore, we argue that while domain specialization can improve robustness in specific cases, it is not a reliable predictor of robustness gains or losses relative to strong GP baselines.

We can also observe from Table~\ref{tab:sensitivity_analysis} that robustness depends not only on model type, but also on model scale and adaptation quality. Smaller models generally show greater sensitivity and less consistent behaviour, suggesting that limited capacity may hinder stable decision-making under semantically equivalent reformulations. By contrast, several of the strongest robustness results are concentrated among 7B--8B models, where both GP and DS models more frequently achieve lower MVS and WCI values. However, scale alone does not guarantee robustness, as some smaller models still achieve competitive stability (e.g., Qwen2.5-1.5B with the best GP MedQA WCI of 15.00\%), while some 8B models remain less stable than peers on specific metrics. This indicates that robustness emerges from the interaction of model capacity, architecture, and training quality rather than parameter count alone.

Finally, both GP and DS models become less robust on the more challenging MedQA datasets. Across most models, MedQA provides higher MVS and WCI than DiagnosisQA, indicating that semantic stability degrades as reasoning demands increase. For example, LLaMA3.1-8B rises from 20.20\% WCI on DiagnosisQA to 26.00\% on MedQA, while Qwen3-8B-Biomedical increases sharply from 18.18\% to 37.00\%. At the same time, confidence variation remains comparatively small for many models despite these correctness shifts, implying that models often maintain similar confidence even when their answers change under equivalent reformulations. These results complement our earlier argument that robustness is shaped more by architecture, scale, and training quality than by domain specialization alone.

\begin{figure*}[t]
    \centering
    
\includegraphics[width=0.9\textwidth]{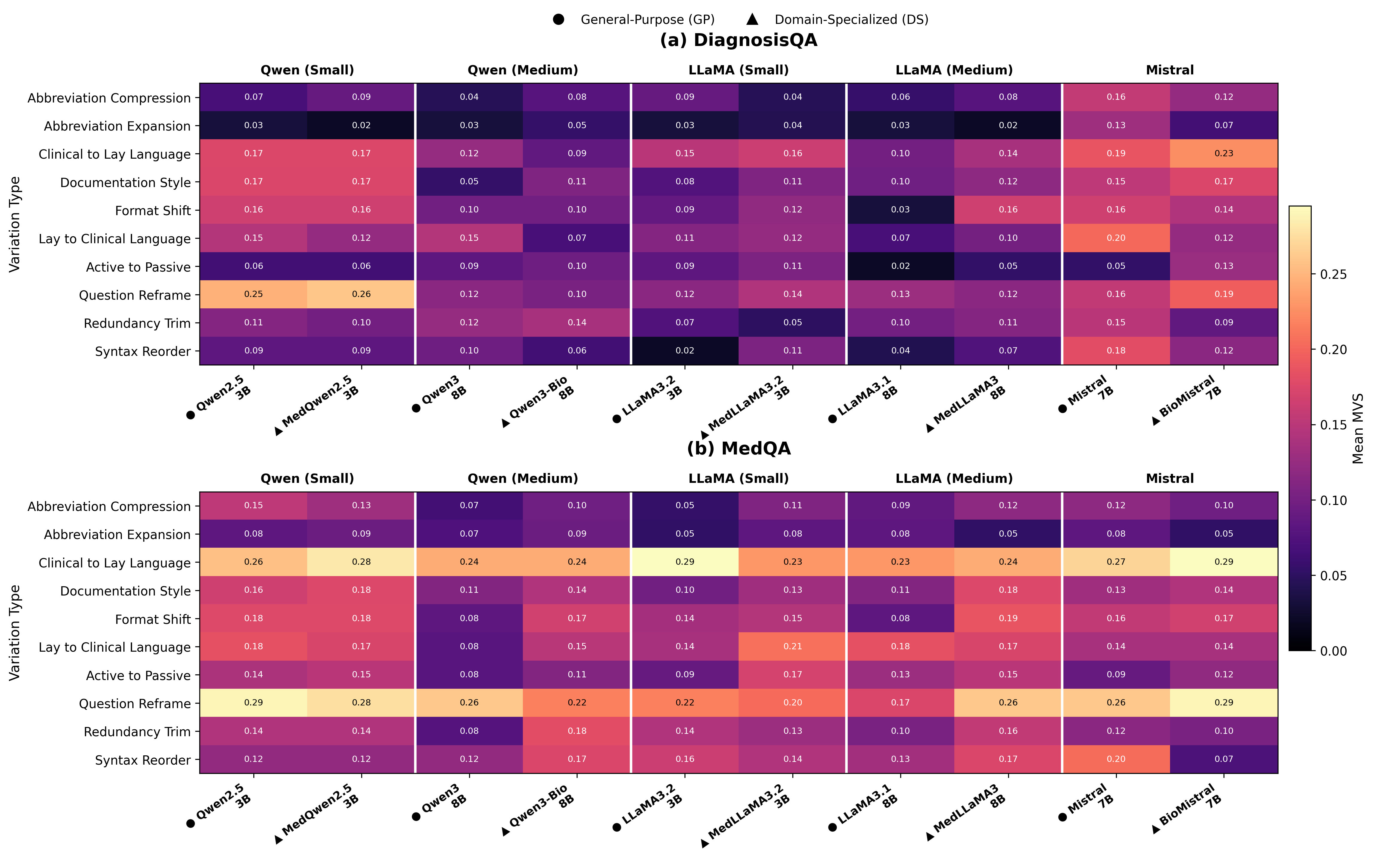}
     
    \caption{Robustness analysis across meaning-preserving variations for GP and DS LLMs. The heatmaps report the mean MVS across different variation types for (a) DiagnosisQA and (b) MedQA datasets. Lower values indicate stronger semantic stability. }
    \label{fig:variations_sensitivity}
\end{figure*}

\begin{figure*}[!ht]
    \centering
    
    \includegraphics[width=0.8\linewidth]{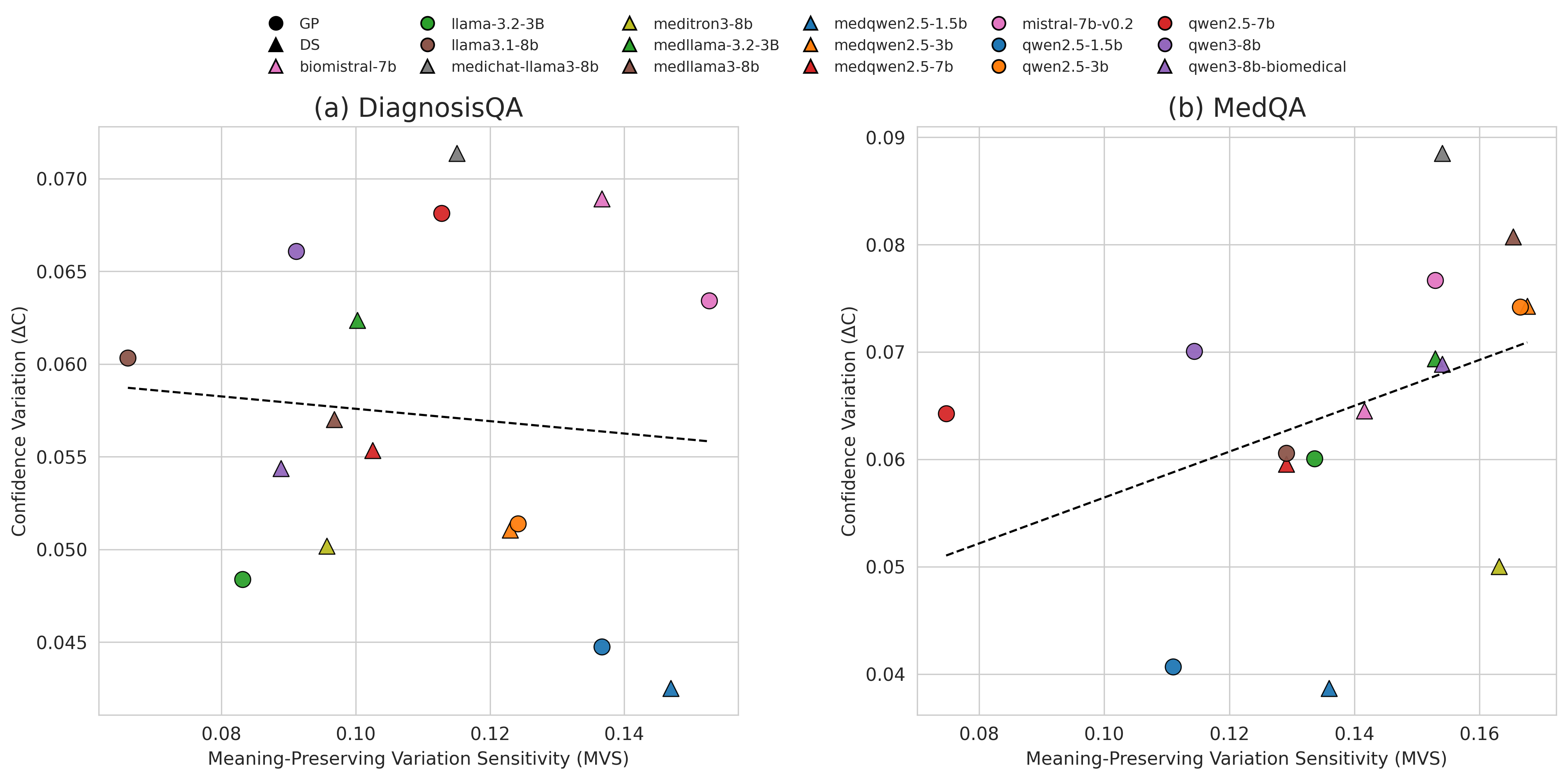}
     
    \caption{Relationship between prediction instability and confidence instability under meaning-preserving prompt reformulations in terms average MVS and $\Delta$C on (a) DiagnosisQA and (b) MedQA. Lower-left regions indicate models with both stable predictions and stable confidence under reformulation, whereas upper-right regions indicate joint instability.}
    \label{fig:scatter}
\end{figure*}

\begin{figure*}[t]
    \centering
    
    \includegraphics[width=0.8\linewidth]{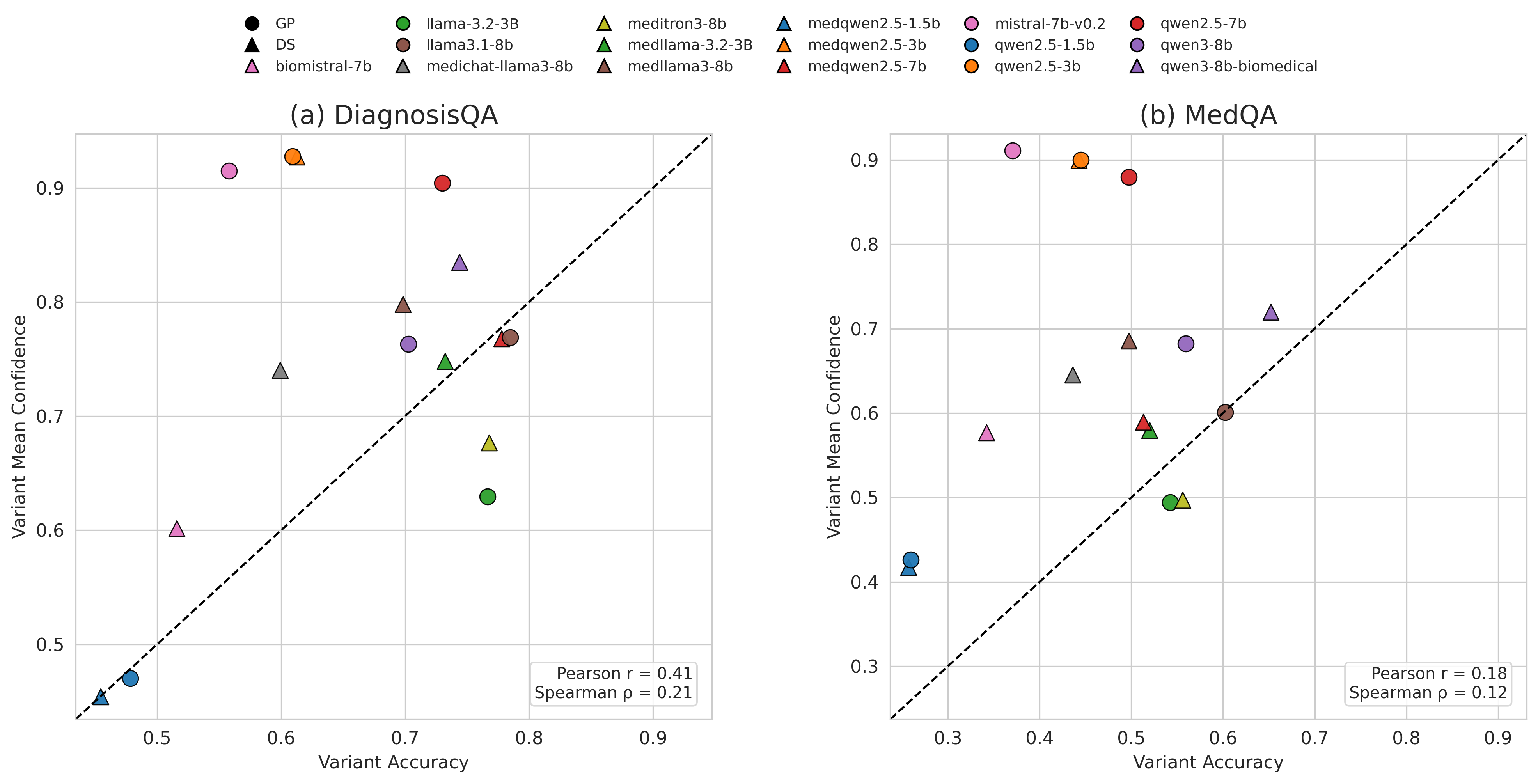}
     
    \caption{Confidence–accuracy alignment across various GP and DS models on DiagnosisQA and MedQA datasets. The dashed line ($y=x$) indicates perfect alignment between confidence and performance. Pearson's $r$ and Spearman's $\rho$ report the confidence–accuracy correlation for each dataset.}
    \label{fig:conf_vs_acc}
\end{figure*}

\subsubsection{Variation-Specific Sensitivity Analysis}
Figure~\ref{fig:variations_sensitivity} provides a paired robustness comparison between GP and DS models under various meaning-preserving linguistic reformulations. Each GP model is aligned with a parameter-matched DS counterpart, allowing robustness differences to be interpreted with reduced confounding from model scale. Two high-level trends are immediately apparent from Figure~\ref{fig:variations_sensitivity}. First, the dataset effect observed in Table~\ref{tab:sensitivity_analysis} is again confirmed: MedQA consistently exhibits higher MVS than DiagnosisQA across nearly all model pairs and transformation types. This suggests that robustness degrades when input tasks demand more complex reasoning, denser factual recall, or greater sensitivity to contextual nuance. Second, robustness remains strongly model-dependent rather than domain-dependent (i.e., GP vs DS), where domain specialization does not uniformly improve stability. In some pairs, DS models outperform their GP counterparts---for example, BioMistral is more stable than Mistral under several MedQA transformations, and Qwen3-Bio generally improves over Qwen3-8B on DiagnosisQA. However, other DS variants show little benefit or mixed differences relative to their paired GP baselines, particularly in the Qwen small-scale pair, where MedQwen2.5-3B often mirrors or only marginally differs from Qwen2.5-3B. These results reinforce that domain specialization neither consistently helps nor harms robustness, with effects varying by architecture, scale, and dataset.

The figure also reveals that robustness is highly dependent on the type of linguistic reformulation. Across both datasets, \textit{Question Reframe} is consistently among the most disruptive transformations and often produces the highest MVS scores across various models. This suggests that models remain sensitive to changes in discourse framing or interrogative structure, even when semantic intent is preserved. \textit{Clinical-to-Lay Language} is similarly challenging, particularly on MedQA, where many models attain sharp increases in sensitivity. This implies that translating between professional and consumer-facing terminology remains a non-trivial robustness challenge, even for medically tuned LLMs. In contrast, variations focusing on localized edits such as \textit{Abbreviation Expansion}, \textit{Abbreviation Compression}, and \textit{Active-to-Passive} voice changes generally lead to lower MVS values, indicating stronger invariance to lexical substitution and shallow syntactic alternation. Another notable pattern is that larger or stronger baseline models are not always the most stable: several 8B models remain vulnerable to question reframing and terminology shifts despite higher nominal accuracy. Overall, the paired comparisons suggest that semantic robustness is shaped by a combination of model architecture, adaptation strategy, and training diversity rather than by parameter scale or medical specialization alone. These findings further emphasize that evaluating medical LLMs solely on standard benchmark accuracy can obscure substantial fragility under realistic variations in clinical language.

\subsection{Confidence Stability Under Meaning-Preserving Reformulations (RQ3)}
We next investigate whether model confidence remains stable when prompts are reformulated without changing their meaning, and whether confidence instability is aligned with prediction instability. As shown in Figure~\ref{fig:scatter}, the association between MVS and $\Delta$C is weak overall, with almost no clear trend on DiagnosisQA (Figure~\ref{fig:scatter}a) and a modest positive trend on MedQA (Figure~\ref{fig:scatter}b). Models that are more prediction-sensitive under reformulation often exhibit larger confidence shifts, yet the relationship is far from deterministic. Several models with relatively modest MVS still show noticeable confidence volatility, whereas others with higher sensitivity maintain comparatively stable confidence. This dispersion indicates that prediction instability and confidence instability capture related but distinct failure modes. A model may preserve the same answer while substantially changing its certainty, or conversely flip predictions while expressing similar confidence. In practical terms, confidence scores alone do not provide a dependable proxy for semantic robustness. A second important finding is that MedQA tends to induce somewhat greater confidence instability than DiagnosisQA. The broader spread of points and steeper trend in Figure~\ref{fig:scatter}b suggest that more challenging clinical reasoning tasks can amplify fluctuations not only in predictions but also in certainty estimates. This implies that confidence becomes less stable precisely in settings where trustworthy model behavior is most critical. Thus, confidence estimates appear sensitive to task difficulty and prompt formulation, rather than reflecting a fixed internal measure of certainty.

To further assess whether model confidence reliably reflects actual performance, we analyzed the alignment between predictive confidence and empirical accuracy across all evaluated models. For each model and dataset, we computed mean variant predictive confidence and compared it against the observed accuracy. The results for this correlation are depicted in Figure \ref{fig:conf_vs_acc}, where the diagonal reference line ($y=x$) denotes perfect alignment between confidence and achieved performance, such that models positioned above the line exhibit confidence levels that exceed their observed accuracy. To quantify the overall association, we additionally report Pearson's $r$ and Spearman's $\rho$ for each dataset. As shown in Figure \ref{fig:conf_vs_acc}, confidence is not consistently aligned with performance across models. On DiagnosisQA, the relationship between confidence and accuracy is modest ($r=0.41$), suggesting that more accurate models tended to be somewhat more confident overall. In contrast, alignment is relatively weaker on MedQA ($r=0.18$), where higher confidence did not meaningfully correspond to higher accuracy. Several models on both benchmarks also appeared above the diagonal line, indicating confidence levels substantially higher than their achieved performance. This suggests that predictive confidence should be interpreted cautiously, as it does not always serve as a dependable indicator of correctness in medical QA tasks.

\subsection{Effect of Quantization on Stability Estimates} Our main sensitivity analysis evaluates all models under 4-bit NF4 quantization with half-precision compute. To quantify the effect of quantization on model sensitivity, we re-evaluated all eight 7--8B models on both DiagnosisQA and MedQA at full precision (16-bit) for paired GP and DS models (i.e., same size and family). The results of these experiments are summarized in Table~\ref{tab:quant-ablation}, which are reported in terms of mean and standard deviation (to facilitate a direct comparison with Table~\ref{tab:sensitivity_analysis}). The table also reports the calculated change ($\Delta$) in the metric introduced by NF4 quantization with respect to full-precision settings, where positive values indicate degradation. The table shows that quantization has a small impact on model stability across both GP and DS models. For MVS, the observed changes are consistently small, with the difference ($\Delta$) for many models very close to zero. However, some models exhibit relatively large degradation in MVS, e.g., LLaMA3.1-8B and MedQwen2.5-7B both move by $+0.04$ on DiagnosisQA, Mistral-7B-v0.2 by $+0.04$ on MedQA, and MedLLaMA3-8B by $+0.06$ on MedQA. This suggests that quantization introduces only negligible additional variance in model stability. A similar pattern is observed for $\Delta$C, where differences remain close to zero for nearly all models, indicating that this metric is largely insensitive to reduced precision. In contrast, WCI exhibits comparatively larger variations, with several models showing increases of up to approximately 8\%, although these effects are not uniform and for Qwen2.5-7B, quantization leads to an improvement of 7\% in WCI, and Qwen3-8B-Biomedical is the only model whose DiagnosisQA WCI improves under quantization ($-1.01$). The asymmetry between datasets is also noticeable for individual models: LLaMA3.1-8B shifts by $+8.08$ on DiagnosisQA but $0.00$ on MedQA, while Qwen2.5-7B moves in opposite directions on the two datasets ($+6.06$ vs $-7.00$). These results highlight that NF4 quantization preserves models' stability to a large extent, with only modest and metric- and dataset-dependent variations.

\begin{table}[!t]
\centering
\caption{Effect of 4-bit NF4 quantization on stability for the eight models. The FP column reports the Full-Precision result as mean$\pm$std; the $\Delta$ column reports the change in the mean under NF4 quantization (NF4$-$FP), so positive values indicate degradation. NF4 values with full standard deviations are reported in Table~\ref{tab:sensitivity_analysis}. Bolded $\Delta$ entries highlight the largest absolute shifts within each metric column.}
\label{tab:quant-ablation}
\footnotesize
\resizebox{\columnwidth}{!}{%
\begin{tabular}{l cc cc cc}
\toprule
& \multicolumn{2}{c}{\textbf{MVS} $(\downarrow)$} & \multicolumn{2}{c}{\textbf{$\Delta C$} $(\downarrow)$} & \multicolumn{2}{c}{\textbf{WCI (\%)} $(\downarrow)$} \\
\cmidrule(lr){2-3} \cmidrule(lr){4-5} \cmidrule(lr){6-7}
\textbf{Model} & FP & $\Delta$ & FP & $\Delta$ & FP & $\Delta$ \\
\midrule
\multicolumn{7}{c}{\textit{DiagnosisQA}} \\
\midrule
\multicolumn{7}{l}{\textbf{General-Purpose (GP) Models}} \\
Qwen3-8B            & $0.09 \pm 0.18$ & $+0.01$            & $0.06 \pm 0.05$ & $+0.01$            & $23.23$ & $+5.05$           \\
Qwen2.5-7B          & $0.10 \pm 0.23$ & $+0.01$            & $0.05 \pm 0.07$ & $\mathbf{+0.02}$   & $24.24$ & $\mathbf{+6.06}$  \\
LLaMA3.1-8B         & $0.03 \pm 0.10$ & $\mathbf{+0.04}$   & $0.05 \pm 0.04$ & $+0.01$            & $12.12$ & $\mathbf{+8.08}$  \\
Mistral-7B-v0.2     & $0.13 \pm 0.23$ & $+0.03$            & $0.06 \pm 0.08$ & $\hphantom{+}0.00$ & $23.23$ & $+5.05$           \\
\addlinespace
\multicolumn{7}{l}{\textbf{Domain-Specific (DS) Models}} \\
Qwen3-8B-Biomedical & $0.08 \pm 0.18$ & $+0.01$            & $0.05 \pm 0.06$ & $\hphantom{+}0.00$ & $19.19$ & $-1.01$           \\
MedQwen2.5-7B       & $0.06 \pm 0.16$ & $\mathbf{+0.04}$   & $0.05 \pm 0.04$ & $\hphantom{+}0.00$ & $22.22$ & $\mathbf{+8.08}$  \\
MedLLaMA3-8B        & $0.10 \pm 0.21$ & $-0.01$            & $0.06 \pm 0.05$ & $\hphantom{+}0.00$ & $22.22$ & $+2.02$           \\
BioMistral-7B       & $0.13 \pm 0.20$ & $+0.01$            & $0.06 \pm 0.04$ & $+0.01$            & $31.31$ & $+3.03$           \\
\midrule
\multicolumn{7}{c}{\textit{MedQA}} \\
\midrule
\multicolumn{7}{l}{\textbf{General-Purpose (GP) Models}} \\
Qwen3-8B            & $0.14 \pm 0.22$ & $-0.02$            & $0.07 \pm 0.05$ & $\hphantom{+}0.00$ & $31.00$ & $+1.00$           \\
Qwen2.5-7B          & $0.07 \pm 0.14$ & $\hphantom{+}0.00$ & $0.06 \pm 0.07$ & $+0.01$            & $25.00$ & $\mathbf{-7.00}$  \\
LLaMA3.1-8B         & $0.09 \pm 0.18$ & $+0.03$            & $0.06 \pm 0.04$ & $\hphantom{+}0.00$ & $26.00$ & $\hphantom{+}0.00$ \\
Mistral-7B-v0.2     & $0.11 \pm 0.19$ & $\mathbf{+0.04}$   & $0.06 \pm 0.07$ & $\mathbf{+0.02}$   & $25.00$ & $+2.00$           \\
\addlinespace
\multicolumn{7}{l}{\textbf{Domain-Specific (DS) Models}} \\
Qwen3-8B-Biomedical & $0.12 \pm 0.21$ & $+0.03$            & $0.06 \pm 0.04$ & $+0.01$            & $32.00$ & $+5.00$           \\
MedQwen2.5-7B       & $0.13 \pm 0.19$ & $\hphantom{+}0.00$ & $0.05 \pm 0.04$ & $+0.01$            & $29.00$ & $+3.00$           \\
MedLLaMA3-8B        & $0.10 \pm 0.19$ & $\mathbf{+0.06}$   & $0.08 \pm 0.06$ & $\hphantom{+}0.00$ & $25.00$ & $+4.00$           \\
BioMistral-7B       & $0.12 \pm 0.21$ & $+0.02$            & $0.07 \pm 0.04$ & $-0.01$            & $27.00$ & $+1.00$           \\
\bottomrule
\end{tabular}}
\end{table}

\subsection{Statistical Significance Analysis}
To determine whether domain specialization systematically alters robustness under meaning-preserving prompt reformulations, we perform pairwise statistical comparisons between strictly matched GP and DS models from the same architectural family and comparable parameter scale (e.g., Qwen vs. MedQwen, LLaMA vs. MedLLaMA, Mistral vs. BioMistral). For each matched pair, we compare aligned per-sample sensitivity metrics using two-sided paired permutation tests (100{,}000 permutations). The results of this analysis are summarized in Table~\ref{tab:gp_ds_comparison} in terms of DS--GP differences in MVS and $\Delta$C, $p$-values, and paired effect sizes ($d$). The table reveals a highly heterogeneous pattern, providing little evidence that domain specialization consistently improves robustness. For instance, most matched comparisons are statistically non-significant across both datasets, and observed effect sizes are generally small. On DiagnosisQA, several DS models show numerically lower MVS than their GP counterparts, including BioMistral (7B), MedQwen2.5 (7B), and Qwen3-Biomedical (8B), but none of these differences are statistically significant. Conversely, MedLLaMA3 (8B) and MedLLaMA3.2 (3B) exhibit higher MVS than their matched GP baselines, again without reliable statistical significance evidence. This supports our finding that semantic robustness is largely model-specific rather than determined by specialization status.

A clearer trend emerges on MedQA, where task difficulty is higher and prompt sensitivity becomes more pronounced. Here, the only statistically significant MVS result is the 7B Qwen pair, where MedQwen2.5-7B is significantly less robust than Qwen2.5-7B ($\mathrm{MVS}_{\mathrm{DS-GP}}$ = +0.052, $p=0.0235$, $d=0.230$). Some other DS models also show positive MVS differences, including MedLLaMA3 (8B) and Qwen3-Biomedical (8B), indicating reduced stability relative to their GP counterparts, although these differences are not statistically significant. Importantly, none of the DS models demonstrates a significant MVS advantage on MedQA. These results support our argument that robustness differences are dataset-specific and that domain specialization does not reliably translate into stronger semantic invariance or weaker robustness in general, with effects varying across matched model pairs. Unlike MVS trends, confidence variation ($\Delta$C) presents a different picture, where significant differences are more frequent than for MVS, indicating that domain specialization more consistently changes how confident models respond to reformulated prompts than whether they preserve the same predictions. However, these shifts are not uniform across all models. Some DS models exhibit lower confidence instability (e.g., MedQwen2.5-1.5B on both datasets and Qwen3-Biomedical on DiagnosisQA), whereas others become significantly less stable (e.g., MedLLaMA3.2 on DiagnosisQA and MedLLaMA3 on MedQA). This asymmetry reinforces a key conclusion of our study: confidence behavior and prediction robustness are related but distinct properties, and gains in one do not guarantee gains in the other.

\begin{table}[!t]
\centering
\caption{Comparison of GP and DS models across DiagnosisQA and MedQA. Reported are differences in MVS, $\Delta$C, $p$-values, and paired effect sizes ($d$). Differences are computed as DS--GP. Negative values indicate improved robustness for DS models. Rows are ordered by model family and parameter scale to facilitate within-family comparisons.}
\label{tab:gp_ds_comparison}
\scalebox{0.5}{
\begin{tabular}{lllcccccc}
\toprule
\textbf{Params} & \textbf{GP Model} & \textbf{DS Model} & \textbf{$\mathrm{MVS}_{\mathrm{DS-GP}}$} & \textbf{$p$(MVS)} & \textbf{$d$(MVS)} & \textbf{$\Delta C_{\mathrm{DS-GP}}$} & \textbf{$p$($\Delta$C)} & \textbf{$d$($\Delta$C)} \\
\midrule

\multicolumn{9}{c}{\textbf{DiagnosisQA}} \\
\midrule

1.5B & Qwen2.5 & MedQwen2.5 & 0.009 & 0.5944 & 0.056 & -0.002 & 0.0073$^{*}$ & -0.263 \\
3B   & Qwen2.5 & MedQwen2.5 & -0.001 & 0.8652 & -0.026 & -0.001 & 0.4471 & -0.078 \\
7B   & Qwen2.5 & MedQwen2.5 & -0.013 & 0.6542 & -0.047 & \cellcolor{best}-0.013 & 0.1602 & -0.143 \\
8B   & Qwen3 & Qwen3-Biomedical & -0.010 & 0.6962 & -0.039 & -0.012 & 0.0234$^{*}$ & -0.231 \\

3B   & LLaMA3.2 & MedLLaMA3.2 & \cellcolor{worst}0.022 & 0.3142 & 0.102 & \cellcolor{worst}0.014 & 0.0006$^{*}$ & 0.351 \\
8B   & LLaMA3.1 & MedLLaMA3 & \cellcolor{worst}0.027 & 0.2536 & 0.116 & -0.003 & 0.5639 & -0.058 \\

7B   & Mistral & BioMistral & \cellcolor{best}-0.022 & 0.5185 & -0.065 & 0.005 & 0.5468 & 0.061 \\

\midrule
\multicolumn{9}{c}{\textbf{MedQA}} \\
\midrule

1.5B & Qwen2.5 & MedQwen2.5 & 0.026 & 0.0636 & 0.182 & -0.002 & 0.0003$^{*}$ & -0.358 \\
3B   & Qwen2.5 & MedQwen2.5 & 0.003 & 0.6047 & 0.055 & \cellcolor{worst}0.000 & 0.8015 & 0.026 \\
7B   & Qwen2.5 & MedQwen2.5 & \cellcolor{worst}0.052 & 0.0235 & 0.230 & -0.006 & 0.4199 & -0.081 \\
8B   & Qwen3 & Qwen3-Biomedical & 0.039 & 0.1641 & 0.141 & -0.001 & 0.7694 & -0.029 \\

3B   & LLaMA3.2 & MedLLaMA3.2 & 0.021 & 0.4434 & 0.077 & 0.009 & 0.0341 & 0.213 \\
8B   & LLaMA3.1 & MedLLaMA3 & 0.038 & 0.2821 & 0.109 & \cellcolor{worst}0.019 & 0.0010$^{*}$ & 0.332 \\

7B   & Mistral & BioMistral & \cellcolor{best}-0.004 & 0.8977 & -0.013 & \cellcolor{best}-0.011 & 0.2665 & -0.112 \\

\bottomrule
\end{tabular}}
\end{table}

\section{Limitations and Future Work}
Despite the promising findings, this study has several limitations that motivate future work. First, high-performing clinical models such as MedPaLM and MedPaLM 2 were not included due to access and deployment constraints, limiting comparisons to publicly available or deployable models; incorporating such systems would enable a more comprehensive evaluation across open-source, GP, and proprietary clinical LLMs. Second, our evaluation protocol intentionally employs standardized single-step decoding under constrained inference settings to isolate intrinsic model sensitivity under controlled conditions. Consequently, the study does not evaluate reasoning-oriented inference strategies, such as chain-of-thought prompting, self-consistency sampling, or adaptive test-time reasoning, which are increasingly adopted in modern LLM deployments. Incorporating such mechanisms would introduce additional inference-time adaptation and model-specific reasoning behaviors that could confound the attribution of instability to semantic-preserving prompt perturbations alone. Therefore, the reported results should be interpreted primarily as estimates of intrinsic decoding-level sensitivity rather than robustness after downstream stabilization or reasoning-based mitigation strategies have been applied. Third, our experiments are limited to multiple-choice benchmarks (MedQA-USMLE and DiagnosisQA). Although these datasets enable controlled and reproducible comparisons, they do not fully reflect the complexity of real-world clinical scenarios, which often involve free-text clinical notes, longitudinal patient histories, retrieval-augmented reasoning, and multi-turn interactions. Fourth, verification of meaning-preserving variations relies on a single clinical expert; although conducted under blinded conditions and showing high agreement with the automated pipeline, the absence of multiple annotators limits assessment of inter-rater reliability. In our future work, we will extend this framework to reasoning-mode and frontier models, apply standard prompt engineering and self-consistency techniques as additional baselines, report multiple-testing-adjusted p-values and inter-rater Cohen's kappa, and validate variations against a small set of human-written paraphrases.

\section{Conclusions}
\label{sec:Conclusion}
In this paper, we introduce a semantically grounded method for evaluating the robustness of medical Large Language Models (LLMs) to meaning-preserving prompt variations. Specifically, we put forward a benchmark of semantically equivalent prompts, developed using a multi-stage pipeline that establishes semantic equivalence via bidirectional entailment using multiple Natural Language Inference (NLI) models, further refines candidate variations through two LLM judges, and conducts final auditing by a clinical expert. In addition, we introduce three complementary robustness metrics: (1) Meaning-Preserving Variation Sensitivity (MVS); (2) confidence variation ($\Delta \text{C}$); and (3) Worst-Case Instability (WCI)---to capture prediction inconsistency, confidence shifts, and per-sample fragility under semantically equivalent reformulations. We perform a large-scale analysis of 16 open-source general-purpose (GP) and domain-specific (DS) medical LLMs across matched model families and parameter scales on the proposed benchmark, derived from the DiagnosisQA and MedQA datasets. Our results reveal that robustness to meaning-preserving variations is highly model-dependent, with no consistent advantage for domain-specialized models. In several experiments comparing GP and DS models, GP models were found equally or more robust than their medically adapted counterparts, while in some matched comparisons DS models improved upon their GP counterparts, indicating that specialization effects are architecture- and dataset-dependent rather than uniform. This suggests that medical fine-tuning alone does not necessarily guarantee stable behavior under natural prompt variability. We also observe that confidence estimates are only weakly aligned with prediction stability, with several models maintaining high confidence even when robustness is reduced or predictions are unstable. Our results highlight the importance of moving beyond standard benchmark accuracy when assessing the clinical readiness of LLMs, and demonstrate that robustness to semantically equivalent prompt reformulations should be treated as a core requirement before deployment in safety-critical healthcare environments.

\bibliographystyle{IEEEtran}
\bibliography{refs}

\end{document}